\documentclass[runningheads,a4paper]{llncs}

\usepackage[pdftex]{graphics}
\usepackage[hyphens]{url}
\usepackage{epstopdf}
\usepackage{microtype}
\usepackage{times}
\usepackage{amsmath,bm}
\usepackage{amssymb}
\usepackage{graphicx}
\usepackage{makecell}
\usepackage{lscape}
\usepackage{url}
\usepackage{comment}
\usepackage[utf8]{inputenc}
\usepackage{float}
\usepackage{hyperref}
\usepackage{xcolor}
\usepackage{bm}
\usepackage{soul}
\usepackage{siunitx}
\usepackage{textpos}

\usepackage{tcolorbox}

\newcommand{\keywords}[1]{\par\addvspace\baselineskip
\noindent\keywordname\enspace\ignorespaces#1}

\begin{document}

\title{
Prediction Uncertainty Estimation for Hate Speech Classification}

\titlerunning{Prediction Uncertainty Estimation for Hate Speech Classification}

\author{Kristian Miok$^1$ \and Dong Nguyen-Doan$^1$ \and Bla\v{z} \v{S}krlj$^2$ \and Daniela Zaharie$^1$ \and \\ Marko Robnik-\v{S}ikonja$^3$}


\authorrunning{Kristian Miok et al.}

\institute{$^1$ Computer Science Department, West University of Timisoara,\\
Bulevardul Vasile Pârvan 4, 300223 Timișoara, Romania \\ \email{\{kristian.miok,dong.nguyen10,daniela.zaharie\}@e-uvt.ro}\\
$^2$Jo\v{z}ef Stefan Institute and Jo\v{z}ef Stefan International Postgraduate School,\\
Jamova 39, 1000 Ljubljana, Slovenia\\
\email{blaz.skrlj@ijs.si}\\
$^3$ Faculty of Computer and Information Science, University of Ljubljana,\\ 
Ve\v{c}na pot 113, 1000 Ljubljana, Slovenia\\
\email{marko.robnik@fri.uni-lj.si}\\
}

\index{Miok, Kristian}
\index{Nguyen-Doan, Dong}
\index{\v{S}krlj, Bla\v{z}}
\index{Zaharie, Daniela}
\index{Robnik-\v{S}ikonja, Marko}
\toctitle{} \tocauthor{}

\maketitle
\begin{tcolorbox}
 The final authenticated publication is available online at SLSP 2019 conference link: \url{https://doi.org/10.1007/978-3-030-31372-2_24}.
\end{tcolorbox}

\begin{abstract}
As a result of social network popularity, in recent years, hate speech phenomenon has significantly increased. Due to its harmful effect on minority groups as well as on large communities, there is a pressing need for hate speech detection and filtering. However, automatic approaches shall not jeopardize free speech, so they shall accompany their decisions with explanations and assessment of uncertainty.  Thus, there is a need for predictive machine learning models that not only detect hate speech but also help users understand when texts cross the line and become unacceptable. 

The reliability of predictions is usually not addressed in text classification. We fill this gap by proposing the adaptation of deep neural networks that can efficiently estimate prediction uncertainty. To reliably detect hate speech, we use Monte Carlo dropout regularization, which mimics Bayesian inference within neural networks. We evaluate our approach using different text embedding methods. We visualize the reliability of results with a novel technique that aids in understanding the classification reliability and errors. 

\end{abstract}

\section{Introduction}

Hate speech represents written or oral communication that in any way discredits a person or a group based on characteristics such as race, color, ethnicity, gender, sexual orientation, nationality, or religion \cite{warner2012detecting}. Hate speech targets disadvantaged social groups and harms them both directly and indirectly \cite{waldron2012harm}. Social networks like Twitter and Facebook, where hate speech frequently occurs, receive many critics for not doing enough to deal with it. As the connection between hate speech and the actual hate crimes is high \cite{bleich2011rise},  the importance of detecting and managing hate speech is not questionable. Early identification of users who promote such kind of communication can prevent an escalation from speech to action. However, automatic hate speech detection is difficult, especially when the text does not contain explicit hate speech keywords. Lexical detection methods tend to have low precision because, during classification, they do not take into account the contextual information those messages carry \cite{davidson2017automated}. Recently, contextual word and sentence embedding methods capture semantic and syntactic relation among the words and improve prediction accuracy.

Recent works on combining probabilistic Bayesian inference and neural network methodology attracted much attention in the scientific community \cite{myshkov2016posterior}. The main reason is the ability of probabilistic neural networks to quantify trustworthiness of predicted results. This information can be important, especially in tasks were decision making plays an important role \cite{inproceedings}. The areas which can significantly benefit from prediction uncertainty estimation are text classification tasks which trigger specific actions. Hate speech detection is an example of a task where reliable results are needed to remove harmful contents and possibly ban malicious users without preventing the freedom of speech. In order to assess the uncertainty of the predicted values, the neural networks require a Bayesian framework. 
On the other hand, Srivastava et al. \cite{srivastava2014dropout} proposed a regularization approach, called dropout, which has a considerable impact on the generalization ability of neural networks. The approach drops some randomly selected nodes from the neural network during the training process. Dropout increases the robustness of networks and prevents overfitting. Different variants of dropout improved classification results in various areas \cite{baldi2013understanding}. Gal and Ghahramani \cite{gal2016dropout} exploited the interpretation of dropout as a Bayesian approximation and proposed a Monte Carlo dropout (MCD) approach to estimate the prediction uncertainty. In this paper, we analyze the applicability of Monte Carlo dropout in assessing the predictive uncertainty.

\begin{figure}[b!!]
  \centering
    \includegraphics[width=0.8 \linewidth]{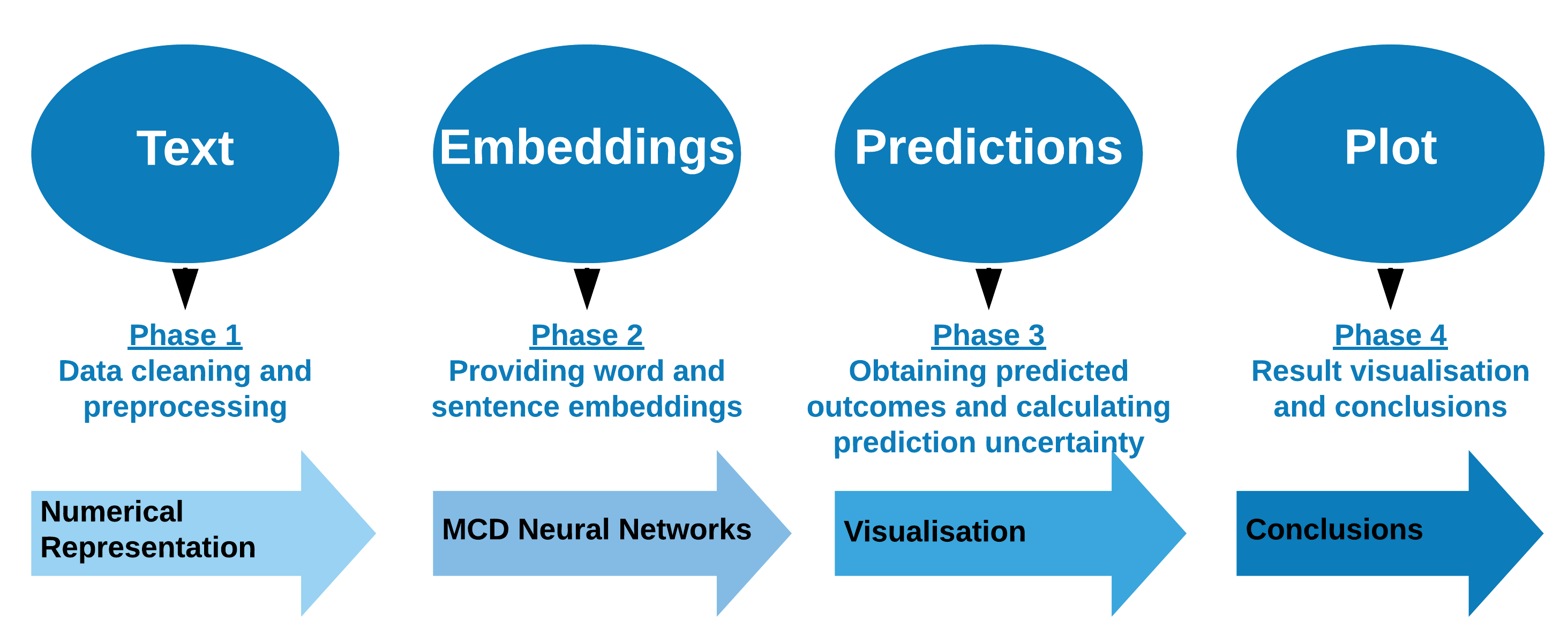}
        \caption{The diagram of the proposed methodology.
        }
        \label{fig:int}
\end{figure}

Our main goal is to accurately and reliably classify different forms of text as hate or non-hate speech, giving a probabilistic assessment of the prediction uncertainty in a comprehensible visual form. We also investigate the ability of deep neural network methods to provide good prediction accuracy on small textual data sets. The outline of the proposed methodology is presented in Figure \ref{fig:int}. 

Our main contributions are: 

\begin{itemize}
    \item investigation of prediction uncertainty assessment to the area of text classification,
    \item implementation of hate speech detection with reliability output,
    \item evaluation of different contextual embedding approaches in the area of hate speech, 
    \item a novel visualization of prediction uncertainty and errors of classification models. 
\end{itemize}
The paper consists of six sections. In Section 2, we present related works on hate speech detection, prediction uncertainty assessment in text classification context, and visualization of uncertainty. In Section 3, we propose the methodology for uncertainty assessment using dropout within neural network models, as well as our novel visualization of prediction uncertainty. Section 4 presents the data sets and the experimental scenario.  We discuss the obtained results in Section 5 and present conclusions and ideas for further work in Section 6.

\section{Related Work}
We shortly present the related work in three areas which constitute the core of our approach: hate speech detection, recurrent neural networks with Monte Carlo dropout for assessment of prediction uncertainty in text classification, and visualization of predictive uncertainty. 

\subsection{Hate Speech Detection}

Techniques used for hate speech detection are mostly based on supervised learning. The most frequently used classifier is the Support Vector Machines (SVM) method \cite{schmidt2017survey}. Recently, deep neural networks,  especially recurrent neural network language models \cite{mehdad2016characters}, became very popular.
Recent studies compare (deep) neural networks \cite{rother2018ulmfit,corazza2018comparing,del2017hate} with the classical machine learning methods.

Our experiments investigate embeddings and neural network architectures that can achieve superior predictive performance to SVM or logistic regression models. More specifically, our interest is to explore the performance of MCD neural networks applied to the hate speech detection task.

\subsection{Prediction Uncertainty in Text Classification}

Recurrent neural networks (RNNs) are a popular choice in text mining. The dropout technique was first introduced to RNNs in 2013 \cite{wang2013fast} but further research revealed negative impact of dropout in RNNs, especially within language modeling. For example, the dropout in RNNs employed on a handwriting recognition task, disrupted the ability of recurrent layers to effectively model sequences \cite{pham2014dropout}. The dropout was successfully  applied to language modeling by \cite{zaremba2014recurrent} who applied it only on fully connected layers. The then state-of-the-art results were explained with the fact that by using the dropout, much deeper neural networks can be constructed without danger of overfitting. 
Gal and Ghahramani \cite{gal2016theoretically} implemented the variational inference based dropout which can also regularize recurrent layers. Additionally, they provide a solution for dropout within word embeddings. The method mimics Bayesian inference by combining probabilistic parameter interpretation and deep RNNs. Authors introduce the idea of augmenting probabilistic RNN models with the prediction uncertainty estimation. Recent works further investigate how to estimate prediction uncertainty within different data frameworks using RNNs \cite{zhu2017deep}. Some of the first investigation of probabilistic properties of SVM prediction is described in the work of Platt \cite{Platt99probabilisticoutputs}. Also, investigation how Bayes by Backprop (BBB) method can be applied to RNNs was done by \cite{fortunato2017bayesian}.  

Our work combines the existing MCD methodology with the latest contextual embedding techniques and applies them to hate speech classification task. The aim is to obtain high quality predictions coupled with reliability scores as means to understand the circumstances of hate speech.

\subsection{Prediction Uncertainty Visualization in Text Classification}

Visualizations help humans in making decisions, e.g., select a driving route, evacuate before a hurricane strikes, or identify optimal methods for allocating business resources. One of the first attempts to obtain and visualize latent space of predicted outcomes was the work of Berger et al. \cite{berger2011uncertainty}. Prediction values were also visualized in geo-spatial research on hurricane tracks \cite{cox2013visualizing,ruginski2016non}. Importance of visualization for prediction uncertainty estimation in the context of decision making was discussed in \cite{liu2019visualizing,liu2016uncertainty}.

We are not aware of any work on prediction uncertainty visualization for text classification or hate speech detection. We present visualization of tweets in a two dimensional latent space that can reveal relationship between analyzed texts.

\section{Deep Learning with Uncertainty Assessment}
\label{sec:MCD}

Deep learning received significant attention in both NLP and other machine learning applications. However, standard deep neural networks do not provide information on reliability of predictions. Bayesian neural network (BNN) methodology can overcome this issue by probabilistic interpretation of model parameters. Apart from prediction uncertainty estimation, BNNs offer robustness to overfitting and can be efficiently trained on small data sets \cite{kucukelbir2017automatic}. However, neural networks that apply Bayesian inference can be computationally expensive, especially the ones with the complex, deep architectures. Our work is based on Monte Carlo Dropout (MCD) method proposed by \cite{gal2016dropout}. The idea of this approach is to capture prediction uncertainty using the dropout as a regularization technique. 

In contrast to classical RNNs, Long Short-term Memory (LSTM) neural networks  introduce additional gates within the neural units. There are two sources of information for specific instance $t$ that flows through all the gates: input values $x_t$ and recurrent values that come from the previous instance $h_{t-1}$. Initial attempts to introduce dropout within the recurrent connections were not successful, reporting that dropout brakes the correlation among the input values. Gal and Ghahramani \cite{gal2016theoretically} solve this issue using predefined dropout mask which is the same at each time step. This opens the possibility to perform dropout during each forward pass through the LSTM network, estimating the whole distribution for each of the parameters. Parameters' posterior distributions that are approximated with such a network structure, $q(\omega)$, is used in constructing posterior predictive distribution of new instances $y^*$:     
\begin{equation}
     p (y^*|x^*,D) \approx \int p\big (y^*|f^\omega(x^*)\big) \; q(\omega) d\omega, 
\end{equation}
where $p\big (y^*|f^\omega(x^*)\big)$ denotes the  likelihood function. In the regression tasks, this probability is summarized by reporting the means and standard deviations while for classification tasks the mean probability is calculated as:

\begin{equation}
    \dfrac{1}{K}\sum_{k=1}^K p(y^*|x^*,\hat{\omega}_k)
\end{equation}
where $\hat{\omega}_k$ $\sim$ $q(\omega)$. Thus, collecting information in $K$ dropout passes throughout the network during the training phase is used in the testing phase to generate (sample) $K$ predicted values for each of the test instance. The benefit of such results is not only to obtain more accurate prediction estimations but also the possibility to visualize the test instances within the generated outcome space.

\subsection{Prediction Uncertainty Visualization}
For each test instance, the neural network  outputs a vector of probability estimates corresponding to the samples generated through Monte Carlo dropout. This creates an opportunity to visualize the variability of individual predictions. With the proposed visualization, we show the correctness and reliability of individual predictions,  including false positive results that can be just as informative as correctly predicted ones. The creation of visualizations consists of the following five steps, elaborated below. 

\begin{enumerate}
    \item Projection of the vector of probability estimates into a two dimensional vector space.
    \item Point coloring according to the mean probabilities computed by the network.
    \item Determining point shapes based on correctness of individual predictions (four possible shapes).
    \item Labeling points with respect to individual documents.
    \item Kernel density estimation of the projected space --- this step attempts to summarize the instance-level samples obtained by the MCD neural network.
\end{enumerate}
As the MCD neural network produces hundreds of probability samples for each target instance, it is not feasible to directly visualize such a multi-dimensional space. To solve this, we leverage the recently introduced UMAP algorithm \cite{mcinnes2018umap}, which projects the input $d$ dimensional data into a $s$-dimensional (in our case $s=2$) representation by using computational insights from the manifold theory.  The result of this step is a two dimensional matrix, where each of the two dimensions represents a latent dimension into which the input samples were projected, and each row represents a text document.

In the next step, we overlay the obtained representation with other relevant information, obtained during sampling. Individual points (documents) are assigned the mean probabilities of samples, thus representing the reliability of individual predictions. We discretize the $[0,1]$ probability interval into four bins of equal size for readability purposes. Next, we shape individual points according to the correctness of predictions. We take into account four possible outcomes (TP - true positives, FP - false positives, TN - true negatives, FN - false negatives).

As the obtained two dimensional projection represents an approximation of the initial sample space, we compute the kernel density estimation in this subspace and thereby outline the main neural network's predictions. We use  two dimensional Gaussian kernels for this task.

The obtained estimations are plotted alongside individual predictions and represent densities of the neural network's focus, which can be inspected from the point of view of correctness and reliability.

\section{Experimental Setting}
We first present the data sets used for the evaluation of the proposed approach, followed by the experimental scenario. The results are presented in Section \ref{sec:Results}.

\subsection{Hate Speech Data Sets}
We use three data sets related to the hate speech.

\subsubsection{1 - HatEval}  data set is taken from the SemEval task "Multilingual detection of hate speech against immigrants and women in Twitter (hatEval)\footnote{\href{https://competitions.codalab.org/competitions/19935}{https://competitions.codalab.org/competitions/19935}}". The competition was organized for two languages, Spanish and English; we only processed the English data set. The data set consists of 100 tweets labeled as 1 (hate speech) or 0 (not hate speech).

\subsubsection{2 - YouToxic} data set is a manually labeled text toxicity data, originally containing \SI{1000} comments crawled from YouTube videos about the Ferguson unrest in 2014\footnote{\href{https://zenodo.org/record/2586669\#.XJiS8ChKi70}{https://zenodo.org/record/2586669\#.XJiS8ChKi70}}. Apart from the main label describing if the comment is hate speech, there are several other labels characterizing each comment, e.g., if it is a threat, provocative, racist, sexist, etc. (not used in our study). There are 138 comments labeled as a hate speech and 862 as non-hate speech. We  produced a data set of 300 comments using all 138 hate speech comments and randomly sampled 162 non-hate speech comments.

\subsubsection{3 - OffensiveTweets}
data set\footnote{\href{https://github.com/t-davidson/hate-speech-and-offensive-language}{https://github.com/t-davidson/hate-speech-and-offensive-language}}  originates in a study regarding hate speech detection and the problem of offensive language \cite{davidson2017automated}. Our data set consists of \SI{3000} tweets. We took 1430 tweets labeled as hate speech and randomly sampled  1670 tweets from the collection of remaining \SI{23353} tweets.

\subsubsection{Data Preprocessing}
Social media text use specific language and contain syntactic and grammar errors. Hence, in order to get correct and clean text data we applied different prepossessing techniques without removing text documents based on the length. The pipeline for cleaning the data was as follows: 
\begin{itemize}
    \item Noise removal: user-names, email address, multiple dots, and hyper-links are considered irrelevant and are removed. \item Common typos are corrected and typical contractions and hash-tags are expanded.
    \item Stop words are removed and the words are lemmatized.
    \end{itemize}

\subsection{Experimental Scenario}

We use logistic regression (LR) and Support Vector Machines (SVM) from the  scikit-learn library \cite{sklearn_api} as the baseline  classification models. As a baseline RNN, the LSTM network from the Keras library was applied \cite{chollet2015keras}. Both LSTM and MCD LSTM networks consist of an embedding layer, LSTM layer, and a fully connected layer within the Word2Vec and ELMo embeddings. The embedding layer was not used in TF-IDF and Universal Sentence encoding.

To tune the parameters of LR (i.e. \textit{liblinear} and \textit{lbfgs} for the solver functions and the number of component $C$ from $0.01$ to $100$) and SVM (i.e. the \textit{rbf} for the kernel function, the number of components $C$ from $0.01$ to $100$ and the gamma $\gamma$ values from $0.01$ to $100$), we utilized the random search approach \cite{bergstra2012random} implemented in scikit-learn. In order to obtain best architectures for the LSTM and MCD LSTM models, various number of units, batch size, dropout rates and so on were  fine-tuned. 

\section{Evaluation and Results}
\label{sec:Results}
We first describe experiments comparing different word representations, followed by sentence embeddings, and finally the visualization of predictive uncertainty.

\subsection{Word Embedding}
In the first set of experiments, we represented the text with word embeddings (sparse TF-IDF \cite{sparck1972statistical} or dense word2vec \cite{mikolov2013efficient}, and ELMo \cite{peters2018deep}). We utilise the gensim library \cite{rehurek_lrec} for word2vec model, the scikit-learn for TFIDF, and the ELMo pretrained model from TensorFlow Hub\footnote{https://tfhub.dev/google/elmo/2}. We compared different classification models using these word embeddings. The results are presented in Table \ref{tab3}.

The architecture of LSTM and MCD LSTM neural networks contains an embedding layer, LSTM layer, and fully-connected layer (i.e. dense layer) for word2vec and ELMo word embeddings. In LSTM, the recurrent dropout is applied to the units for linear transformation of the recurrent state and the classical dropout is used for the units with the linear transformation of the inputs.
The number of units, recurrent dropout, and dropout probabilities for LSTM layer were obtained by fine-tuning (i.e. we used $512$, $0.2$ and $0.5$ for word2vec and TF-IDF, $1024$, $0.5$, and $0.2$ for ELMo in the experiments with MCD LSTM architecture). The search ranges for hyper parameter tuning are described in Table \ref{table:nn_hyperparameters}.

\begin{table}[H]
\footnotesize
	\centering
	\caption{Comparison of classification accuracy (with standard deviation in brackets) for word embeddings, computed using 5-fold cross-validation. All the results are expressed in percentages and the best ones for each data set are in bold.}
	\renewcommand{\arraystretch}{1}
	\setlength{\tabcolsep}{3pt}
	\resizebox{\textwidth}{!}{
	\begin{tabular}{l|rrr|rrr|rrr}
	&\multicolumn{3}{c|}{ \textbf{HatEval}} & \multicolumn{3}{c|}{\textbf{YouToxic}} & \multicolumn{3}{c}{\textbf{OffensiveTweets}}\\	
	 \textbf{Model} & \textbf{TF-IDF} & \textbf{W2V} & \textbf{ELMo}  & \textbf{TF-IDF} & \textbf{W2V} & \textbf{ELMo} & \textbf{TF-IDF} & \textbf{W2V} & \textbf{ELMo}  \\
		\hline		
\textbf{Logistic Regression} & 68.0 [2.4] & 54.0 [13.6] & 62.0 [6.8]&  69.3 [3.0]  & 54.0 [3.0] & \textbf{76.6 [6.1]} & \textbf{77.2 [1.1]} & 68.0 [2.4] & 75.6 [1.2] \\ 
\textbf{SVM} & 63.0 [5.1]& 66.0 [3.7] & 62.0 [12.9] & 70.6 [4.2]  & 55.0 [3.4]  & 73.3 [5.5] & 77.0 [0.7] & 59.6 [1.5] & 73.0 [1.9]\\ 
\textbf{LSTM} & 69.0 [7.3] & 67.0 [6.8]& 66.0 [12.4] & 66.6 [2.3]  & 59.3 [4.6]  & 74.3 [2.7] & 73.4 [0.8] & 75.0 [1.7] & 74.7 [1.9] \\
\textbf{MCD LSTM} & 67.0 [10.8]& \textbf{69.0 [6.6]} & 67.0 [9.8] & 66.0 [3.7]  & 59.3 [3.8]  & 75.3 [5.5]  & 71.1 [1.6] & 72.0 [1.6] & 75.2 [0.9]  \\ 		\hline
	\end{tabular}
	}
	\label{tab3}
\end{table}

\begin{table}[htbp]
\caption{Hyper-parameters for LSTM and MCD LSTM models}
\renewcommand{\arraystretch}{0.75}
\setlength{\tabcolsep}{3pt}
\label{table:nn_hyperparameters}
\centering
\begin{tabular}{l|l|l}
    \textbf{Name} & \textbf{Parameter type} & \textbf{Values}\\
    \hline
    \textbf{Optimizers} & Categorical & Adam, rmsprop\\
    \textbf{Batch size} & Discrete & 4 to 128, step=4\\
    \textbf{Activation function} & Categorical & tanh, relu and linear\\
    \textbf{Number of epochs} & Discrete & 10 to 100, step=5\\
    \textbf{Number of units} & Discrete & 128, 256, 512, or 1024\\
    \textbf{Dropout rate} & Float & 0.1 to 0.8, step=0.05\\
    \hline
\end{tabular}
\end{table}

The classification accuracy for HatEval data set is reported in the Table \ref{tab3} (left). The difference between logistic regression and the two LSTM models indicates accuracy improvement once the recurrent layers are introduced. On the other hand, as the ELMo embedding already uses the LSTM layer to take into account semantic relationship among the words, no notable difference between logistic regression and LSTM models can be observed using this embedding. 

Results for YouToxic and OffensiveTweets data sets are presented in Table \ref{tab3} (middle) and (right), respectively. Similarly to the HatEval data set, there is a difference between the logistic regression and the two LSTM models using the word2vec embeddings. For all data sets, the results with ELMo embeddings are similar across the four classifiers.

\subsection{Sentence Embedding}
In the second set of experiments, we compared different classifiers using sentence embeddings \cite{cer2018universal} as the representation. Table \ref{tab4} (left) displays results for HatEval. We can notice improvements in classification accuracy for all classifiers compared to the word embedding representation in Table \ref{tab3}. The best model for this small data set is MCD LSTM. For larger YouToxic and OffensiveTweets data sets, all the models perform comparably. Apart from the prediction accuracy the four models were compared using precision, recall and F1 score \cite{chi}. 

We use the Universal Sentence Encoder module\footnote{https://tfhub.dev/google/universal-sentence-encoder-large/3} to encode the data. The architecture of LSTM and MCD LSTM contains a LSTM layer and dense layer. With MCD LSTM architecture in the experiments, the number of neurons, recurrent drop\-out and dropout value for LSTM is $1024$, $0.75$ and $0.5$, respectively. The dense layer has the same number of units as LSTM layer, and the applied dropout rate is $0.5$. The hyper-parameters used to tune the LSTM and MCD LSTM models are presented in the Table \ref{table:nn_hyperparameters}.

\begin{table}[H]
	\centering
	\caption{Comparison of predictive models using sentence embeddings. We present average classification accuracy, precision, recall and $F_1$ score (and standard deviations), computed using 5-fold cross-validation. All the results are expressed in percentages and the best accuracies are in bold. 
	} 
	\renewcommand{\arraystretch}{1.2}
	\setlength{\tabcolsep}{4pt}
	\resizebox{\textwidth}{!}{
	\begin{tabular}{l|rrrr|rrrr|rrrr}
	&\multicolumn{4}{c}{ \textbf{HatEval}} & \multicolumn{4}{c}{\textbf{YouToxic}} & \multicolumn{4}{c}{\textbf{OffensiveTweets}}\\
	 \textbf{Model} & \textbf{Accuracy} & \textbf{Precision} & \textbf{Recall} & \textbf{F1} & \textbf{Accuracy} & \textbf{Precision} & \textbf{Recall} & \textbf{F1} & \textbf{Accuracy} & \textbf{Precision} & \textbf{Recall} & \textbf{F1} \\
		\hline	
\textbf{LR} & 66.0 [12.4]  & 67.3 [15.3] & 65.2 [15.9] & 65.2 [13.1] & 77.3  [4.1]& 74.3 [7.3] & 77.3 [3.6] & 75.7 [5.3]  & 80.8 [1.0]& 79.6 [1.9]  & 84.9 [1.2]  & 82.2 [1.1]  \\ 
\textbf{SVM} & 67.0 [12.1] & 68.2 [15.2]  & 65.0 [15.8]  & 65.8 [13.3]  & 77.3 [6.2]& 72.6 [8.6] & 80.7 [7.4] & 76.3 [7.6] & 80.7 [1.3]& 78.6 [2.0]  & 86.7 [1.0]  & 82.4 [1.2]  \\ 
\textbf{LSTM} &  70.0 [8.4]  & 70.8 [11.0]  &  63.1 [17.5]  &  66.2 [14.4]  & 76.6 [8.6] & 73.4 [11.2] & 79.2 [8.0] & 75.8 [8.6] & 80.7 [1.6]& 82.8 [2.1] & 79.7 [2.3] & 81.1 [1.5]   \\	
\textbf{MCD LSTM} & \textbf{74.0 [10.7]} &  73.4 [12.7] &  78.4 [13.6] &  74.9 [10.0] & \textbf{78.7 [5.8]} & 74.7 [9.2]  & 80.9 [6.5] & 77.5 [7.4] & \textbf{81.0 [1.2]} & 81.5 [1.8]  & 82.5 [2.7]  & 81.9 [1.3]   \\
		\hline
	\end{tabular}}
	\label{tab4}
\end{table}

\subsection{Visualizing Predictive Uncertainty}
\begin{figure}[H]
  \centering
    \includegraphics[width=0.5 \linewidth,height=4.5cm]{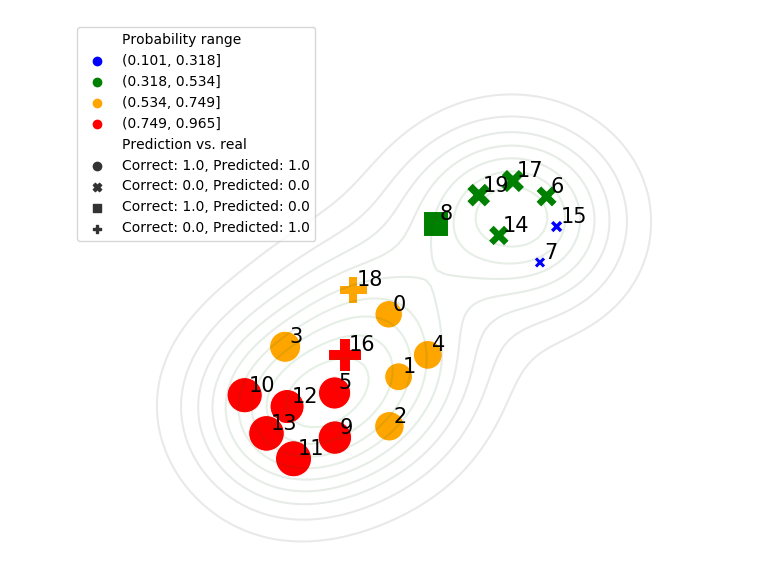}
        \caption{Visualization of individual predictions along with their probability ranges. The numeric labels besides each point correspond to individual tweets, summarized in Table ~\ref{tbl:tweets}. The size of the points corresponds to the neural network's mean probability for a given prediction. True positives are marked with circles, true negatives with crosses, false positives with squares, and false negatives as pluses.
        }
        \label{fig:viz}
\end{figure}

In Figure~\ref{fig:viz} we present a new way of visualizing dependencies among the test tweets. The relations are result of applaing the MCD LSTM network to the HetEval data set. This allows further inspection of the results as well as interpretation of correct and incorrect predictions. To improve comprehensibility of predictions and errors, each point in the visualization is labeled with a unique identifier, making the point tractable to the original document, given in Table~\ref{tbl:tweets}.

\begin{table}[hbt]
\centering
 \caption{Test documents (one fold) used for the visualization in Figure \ref{fig:viz}.}
\resizebox{1\textwidth}{!}{
\begin{tabular}{l}
Index (see Figure \ref{fig:viz}) | Corresponding text \\
     0 |                                                                                                                                                                                   ok if you fu***** said leave block me. but dm me first i am gonna kick your ass. shut the f*** up you  \\
     1 |  wow mo c*** got hard. want to pull your panties down, push you against these door and let you feel my c*** deep in your wet p****  \\
     2 |                                                                                                                                                                                                                  no i said hope. i hope you women learn your place! sitdowninthekitchen \\
     3 |                                                                                                                                                                                     so this b**** blocked me but check her pinned tweet out, she is accusing men of rape on tumblr with \dots  \\
     4 |                                                                                                                                                                             this is not rape, or anything like it. this a crude pass. this is not 'rape culture.' it is w**** culture.  \\
     5 |                                                                                                                                                                                rape is bad but alota h*** be lying bout that s*** women are crazy dawg \& will f*** ya life up be careful \\
     6 |                                                                                                                                                                                                       lol. they did not applaud a rape threat .how long are you going to ride that s***? \\
     7 |                                                                                                                                                                                       you had a great chance to show everyone that youre not a c*** on tv tonight and you f***** it up. \\
     8 |  lmao!!! stuff i say. some stuff be so damn funny.especially when it comes to illegal immigrants who are full 
     fledged criminals coming to america \dots \\
     9 |                                                                                                                                                                                                                the german government pays for 3 week vacation for refugees to go home -  \\
    10 |                                                                                                                                                                                                                                                      burundian refugees should go home!  \\
    11 |                                                                                                                                                                                                no benefits to any immigrants !! come here \&amp; make it on your own, or go back home !!  \\
    12 |                                                                                                                                                                                            illegal aliens waiting for first class msc container-shippping in antwerpen stop immigration  \\
    13 |                                                                                                                sick barstewards! this is what happens when we put up the refugees welcome signs! they not only rape our wives or girlfriends $\dots$  \\
    14 |                                                                                                                  i am not forcing anyone. i am just putting out my view. people believe it or not is their prerogative.even if a single person agree with me $\dots$ \\
    15 |                                                                                                                                                                                                tajikistani charged with beating new york university professor to death another animal./ \\
    16 |                                                                                                                                                                                                            child prostitution: feds break up mexico-to-queens sex trafficking ring via \dots \\
    17 |                                            home office guilty of a serious breach of the duty of candour and cooperation regarding children entitled to enter uk. where did these children go? \dots  \\
    18 |                                                                                                                                                          p.s why do you not pay unemployed people who do endless hours of voluntary work they do that to give something to the community \\
    19 |                                                                                                                                                                                                                       seriously, amy and cindy are bffs, i know that for sure. hmm, mmm. \\
\end{tabular}
}
    \label{tbl:tweets}
\end{table}

As Figure \ref{fig:viz} shows, the tweets are grouped into two clusters. According to the kernel density isometric lines, two centers are identified: the tweets assigned lower probability of being hate speech and the tweets with higher probability of being hate speech.  Let us focus on the wrongly classified tweets and their positions in the graph (tweets 8, 16 and 18). While for tweets 8 and 18 the classifier wasn't certain and a mistake seems possible according to the plot, the tweet 16 was predicted to be hate speech with high probability. Analyzing the words that form this tweet, we notice that not only that most of them often do appear in the hate speech but also this combination of the words used together is very characteristic for the offensive language.

Our short demonstration shows the utility of the proposed visualization which can identify different types of errors and helps to explain weaknesses in the classifier or wrongly labeled data.

\section{Conclusions}

We present the first successful approach to assessment of prediction uncertainty in hate speech classification. Our approach uses LSTM model with Monte Carlo dropout and shows performance comparable to the best competing approaches using word embeddings and superior performance using sentence embeddings. We demonstrate that reliability of predictions and errors of the models can be comprehensively visualized. Further, our study shows that pretrained sentence embeddings outperform even state-of-the-art contextual word embeddings and can be recommended as a suitable representation for this task. The full Python code is publicly available \footnote{https://github.com/KristianMiok/Hate-Speech-Prediction-Uncertainty}.

As persons spreading hate speech might be banned, penalized, or monitored not to put their threats into actions, prediction uncertainty is an important component of decision making and  can help humans observers avoid false positives and false negatives. Visualization of prediction uncertainty can provide better understanding of the textual context within which the hate speech appear. Plotting the tweets that are incorrectly classified and inspecting them can identify the words that trigger wrong classifications.      

Prediction uncertainty estimation is rarely implemented for text classification and other NLP tasks, hence our future work will go in this direction. A recent emergence of cross-lingual embeddings possibly opens new opportunities to share data sets and models between languages. As evaluation in rare languages is difficult, the assessment of predictive reliability for such problems might be an auxiliary evaluation approach. In this context, we also plan to investigate convolutional neural networks with probabilistic interpretation.  

\subsubsection*{Acknowledgments.}
The work was partially supported by the Slovenian Research Agen\-cy (ARRS) core research programme P6-0411. This project has also received funding from the European Union’s Horizon 2020 research and innovation programme under grant agreement No 825153 (EMBEDDIA).

\bibliographystyle{splncs03}
\bibliography{paper}

\end{document}